%% file: main.tex
\documentclass[sigconf,authorversion,nonacm]{acmart}

\renewcommand\footnotetextcopyrightpermission[1]{} 
\usepackage{tikz}
\usepackage{balance}
\usepackage{amsmath}
\usepackage[frozencache=true,cachedir=minted-cache]{minted} 

\usepackage{multirow}
\usepackage{listings}
\usepackage{caption}
\usepackage{subfig}
\usepackage{xcolor}
\usepackage{circledsteps}
\usepackage{comment}
\usepackage{enumitem}
\usepackage{nicefrac}
\usepackage{arydshln}
\usepackage{graphicx}

\usepackage[normalem]{ulem}

\DeclareFixedFont{\ttb}{T1}{txtt}{bx}{n}{6} 
\DeclareFixedFont{\ttm}{T1}{txtt}{m}{n}{12}  
\definecolor{deepblue}{rgb}{0,0,0.5}

\makeatletter
\let\old@lstKV@SwitchCases\lstKV@SwitchCases
\def\lstKV@SwitchCases#1#2#3{}
\makeatother
\usepackage{lstlinebgrd}
\makeatletter
\let\lstKV@SwitchCases\old@lstKV@SwitchCases

\lst@Key{numbers}{none}{%
    \def\lst@PlaceNumber{\lst@linebgrd}%
    \lstKV@SwitchCases{#1}%
    {none:\\%
     left:\def\lst@PlaceNumber{\llap{\normalfont
                \lst@numberstyle{\thelstnumber}\kern\lst@numbersep}\lst@linebgrd}\\%
     right:\def\lst@PlaceNumber{\rlap{\normalfont
                \kern\linewidth \kern\lst@numbersep
                \lst@numberstyle{\thelstnumber}}\lst@linebgrd}%
    }{\PackageError{Listings}{Numbers #1 unknown}\@ehc}}
\makeatother

\newcommand{\xzcomment}[1]{\ignorespaces}
\usepackage{tikz}
\def\checkmark{\tikz\fill[scale=0.4](0,.35) -- (.25,0) -- (1,.7) -- (.25,.15) -- cycle;}

\begin{document}

\title{TorchBench: Benchmarking PyTorch with High API Surface Coverage}

\date{}

\settopmatter{printfolios=true}

\author{Yueming Hao}
\email{yhao24@ncsu.edu}
\affiliation{%
  \institution{North Carolina State University}
  \city{Raleigh}
  \state{North Carolina}
  \country{USA}
}
\author{Xu Zhao}
\email{xzhao9@meta.com}
\affiliation{
  \institution{Meta Platforms, Inc.}
  \city{Menlo Park}
  \state{California}
  \country{USA}
}
\author{Bin Bao}
\email{binbao@meta.com}
\affiliation{
  \institution{Meta Platforms, Inc.}
  \city{Menlo Park}
  \state{California}
  \country{USA}
}
\author{David Berard}
\email{dberard@meta.com}
\affiliation{
  \institution{Meta Platforms, Inc.}
  \city{Menlo Park}
  \state{California}
  \country{USA}
}
\author{Will Constable}
\email{whc@meta.com}
\affiliation{
  \institution{Meta Platforms, Inc.}
  \city{Menlo Park}
  \state{California}
  \country{USA}
}
\author{Adnan Aziz}
\email{adnanaziz@meta.com}
\affiliation{
  \institution{Meta Platforms, Inc.}
  \city{Menlo Park}
  \state{California}
  \country{USA}
}
\author{Xu Liu}
\email{xliu88@ncsu.edu}
\affiliation{%
  \institution{North Carolina State University}
  \city{Raleigh}
  \state{North Carolina}
  \country{USA}
}

\thispagestyle{empty}

\newcommand{\tool}[0]{\mbox{\textsc{TorchBench}}}

\begin{abstract}

Deep learning (DL) has been a revolutionary technique in various domains. To facilitate the model development and deployment, many deep learning frameworks are proposed, among which PyTorch is one of the most popular solutions. The performance of eco-system around PyTorch is critically important, which saves the costs of training models and reduces the response time of model inferences. In this paper, we propose \tool{}, a novel benchmark suite to study the performance of PyTorch software stack. Unlike existing benchmark suites, \tool{} encloses many representative models, covering a large PyTorch API surface. \tool{} is able to comprehensively characterize the performance of the PyTorch software stack, guiding the performance optimization across models, PyTorch framework, and GPU libraries. We show two practical use cases of \tool{}. (1) We profile \tool{} to identify GPU performance inefficiencies in PyTorch. We are able to optimize many performance bugs and upstream patches to the official PyTorch repository. (2) We integrate \tool{} into PyTorch continuous integration system. We are able to identify performance regression in multiple daily code checkins to prevent PyTorch repository from introducing performance bugs.  \tool{} is open source and keeps evolving.

\end{abstract}

\maketitle

\section{Introduction}

Deep learning (DL) has been the most transformative technology over the past decade, given its wide applicability in global climate projections~\cite{haggag2021deep,el2022global}, image processing~\cite{resnext,resnet,mobilenetv3}, speech recognition~\cite{speech1,speech2}, content recommendation~\cite{content1,content2,content3}, and pattern classification~\cite{pattern1,pattern2}, and great potential in emerging domains such as autonomous driving~\cite{rao2018deep,milz2018visual}, augmented/virtual reality~\cite{zhou2020intuitive,karacsony2019brain}, and sciences~\cite{ruff2021alphafold,ramsundar2019deep}. 
To facilitate the development and deployment of DL models,
DL framework practitioners have proposed a number of DL frameworks, such as  Caffe~\cite{jia2014caffe}, TVM~\cite{chen2018tvm}, ONNX~\cite{onnx}, MXNet~\cite{chen2015mxnet}, JAX~\cite{jax2018github}, PyTorch~\cite{pytorch}, and TensorFlow~\cite{tensorflow2015-whitepaper}. These frameworks provide rich APIs and support performance tuning. 

PyTorch is one of the most popular deep learning frameworks. As an open-source project, PyTorch has more than 2000 contributors and has received more than 61,000 stars by the end of 2022. PyTorch is still actively evolving; PyTorch has over 110k commits in 2022~\cite{pytorch_statistics}. 
The machine learning research papers based on PyTorch have increased from 29\% to 65\% in the past four years~\cite{paperswithcode}. 

One major challenge of deep learning tasks is the extensive amount of computation for both training and inference, which translates to huge time and money costs. For example, the best version of AlphaGo~\cite{alphago} needs weeks to train with 280 GPUs, and its estimated cost is 35 million dollars~\cite{greenai}. In addition, the real time inference of deep learning models on edge devices requires fast response and low latency. Thus, it is critically important for PyTorch to achieve bare-metal performance for boosting productivity and reducing costs.

However, understanding and optimizing PyTorch performance is challenging because of PyTorch's complex and rapidly evolved code bases. On the one hand, the PyTorch software stack consists of three major components: the acceleration libraries (e.g., CuDNN~\cite{chetlur2014cudnn}), the PyTorch framework~\cite{pytorch} (including optimizers and compilers), and deep learning models; performance inefficiencies can exist in or across any components. Without a systematic evaluation, inefficiencies can be difficult to identify. On the other hand, as an active open-source project, PyTorch receives many patches for new functionality or performance improvement. Without a thorough evaluation, it is difficult to understand how these patches affect the entire PyTorch software stack for various models in different domains.

Benchmarking is a well-known technique to understand the performance of different workloads, programming languages, compilers, and architectures. For example, SPEC provides many standard benchmark suites for various purposes; Renassance~\cite{prokopec2019renaissance} is a Java benchmark suite; DeathStar~\cite{deathstar} is a benchmark suite for microservices; NPB~\cite{bailey1991parallel} is a benchmark suite for different parallel models in scientific computing; and Rodinia~\cite{che2009rodinia} is a GPU benchmark suite. These benchmarks provide indispensable performance insights for both software and hardware evolution. 
MLPerf~\cite{mattson2020mlperf} is the state-of-the-art benchmark suite for deep learning workloads.
However, existing deep learning benchmark suites, including MLPerf, aim to compare the performance of deep learning models running on different hardware and frameworks. They usually include a small number of deep learning models (e.g., MLPerf has eight models only) and cover a small PyTorch API surface, which fails to identify many PyTorch performance bugs or fairly evaluate the performance impact of patches.

\subsection{Motivating Examples}\label{section:motivation}

We show two examples to motivate the necessity of a comprehensive benchmark suite for PyTorch.

\paragraph{Misleading performance characterization}
Understanding the performance difference across various architectures is one of the major tasks for benchmarking. However, without a comprehensive benchmark suite, such characterization can result in misleading conclusions. For example, some studies~\cite{yin2021comparative} show that AMD GPUs outperform NVIDIA GPUs on PyTorch, and some studies~\cite{wang2020benchmarking, otterness2020amd} show an opposite conclusion. Thus, a comprehensive benchmark suite is necessary for performance characterization. Unlike existing studies, we show a different conclusion and unique insights in Section~\ref{section: amd vs nvidia}.

\sloppy
\paragraph{Missing performance bugs}
Performance bugs can be buried deep in complex PyTorch code bases, especially in the cold execution paths. We identify a recent performance bug in PyTorch, which is inappropriate error handling. 
PyTorch implements an error handling mechanism named c10\_Exception to deal with runtime errors. It prints out backtraces for all errors and uses \texttt{std::string} to generate error messages. Since error handling is usually in the cold path, it does not incur any performance degradation in models of existing benchmark suites.  However, we find that this error code handling slows down quantized models by 10$\times$. Our further investigation shows that quantized models heavily call \texttt{torch.ops} API, which frequently throws a 
benign error, ``Not Implemented Functions'', resulting in a large overhead in error handling. This issue has been confirmed by PyTorch team, and {\tt c10\_Exception} has been reverted to the previous implementation. Thus, a benchmark suite that covers a large PyTorch API surface is necessary to expose performance bugs.

\subsection{Paper Contribution}
In this paper, we develop \tool{}, a novel benchmark suite for PyTorch to address the aforementioned challenges. Unlike existing approaches, \tool{} embraces a large number of models, which cover a large PyTorch API surface.
Given the voluminous deep learning models, we carefully include the representative models in \tool{} for both fairness and generality.
Additionally, we develop a set of tools associated with \tool{} to enable \tool{} to (1) run benchmarks with different configurations, (2) collect various performance statistics, and (3) be ready for any continuous integration systems. \tool{} helps fix many 
performance bugs in PyTorch eco-system, and many of them are accepted by the official PyTorch repository.
In summary, \tool{} makes the following contributions.

\begin{itemize}[leftmargin=*]
    \item \tool{} is the first PyTorch benchmark suite that consists of rich models in different domains. It covers 2.3$\times$ more PyTorch API surface, compared to the state-of-the-art MLPerf benchmark suite.
    
    \item \tool{} integrate a set of built-in tools, which configure execution environments and collect performance statistics. \tool{} is able to report multiple metrics to thoroughly characterize PyTorch performance.
    
    \item \tool{} demonstrates two use cases in practice. First, \tool{} exposes the performance bugs in PyTorch software stacks and provides insights for bug fixing. Second, \tool{} is configured as the continuous integration of PyTorch repository to understand the performance regression of daily committed patches.
\end{itemize}

The rest of the paper is organized as follows. Section~\ref{section: benchmark suite} describes the models, benchmark adaptation, and configurations in our benchmark suite. 
Section~\ref{characterization} characterizes TorchBench from different aspects, including execution time breakdown, different PyTorch compilers, and different GPU platforms. 
Section~\ref{section: applying in practice} shows the practical usages of \tool{} in detecting performance inefficiencies in PyTorch software stack.
Section~\ref{section:related work} reviews related works and distinguishes our approach. Section~\ref{section:conclusion} presents some conclusions.

\section{\tool{} Suite}\label{section: benchmark suite}

\input{table_all_models_4.tex}

There are a huge amount of deep learning models developed by the PyTorch community. For example, The Hugging Face platform has more than 10,000 models~\cite{wolf-etal-2020-transformers}. Thus, it is challenging for \tool{} to enclose representative models to cover performance behaviors of typical PyTorch use cases. We work closely with machine learning engineers and set the following criteria to select models for \tool{}.
\begin{itemize}[leftmargin=*]

\item Classic models. \tool{} includes models that were developed many years ago but have been proven both useful and impactful, such as ResNet~\cite{resnet}, VGG16~\cite{vgg16}, and MobileNet~\cite{mobilenetv3,sandler2018mobilenetv2}. These models serve as the foundation of many state-of-the-art models.

\item Popular models. \tool{} includes popular models released in recent years. These models attract extensive attention in the community, enabling many academic research papers and applications.  These models include pig2~\cite{dalle2}, T5~\cite{t5}, Yolo~\cite{redmon2018yolov3}, and docTR~\cite{doctr2021}.

\item Important models in the industry.  Industrial companies such as Meta and Google release important models which are used in their products. These models include Detectron2~\cite{wu2019detectron2} and Bert~\cite{devlin2018bert}. 

\item Diverse models. \tool{} includes models from different domains to ensure a fair comparison. Moreover, models of different weight layers and different implementations are included.
\end{itemize}
In the rest of this section, we describe the benchmarks, set the study scope, and distinguish \tool{} from existing approaches.

\subsection{Benchmark Description}
Table~\ref{table:all_models_3} overviews all the benchmarks in \tool{}, which consists of 84 deep learning models and covers six domains. For a given model name, it may consist of a prefix and/or a suffix. The prefix, such as {\tt d2} (Detectron2), {\tt hf} (Hugging Face), and {\tt timm}, means the collection or platform the model is from. 
The suffix means different configurations. Specifically, 
{\tt c4} means using a conv4 backbone; {\tt FPN} means using Feature Pyramid Network backbone; {\tt dc5} means using a conv5 backbone with dilations in conv5; {\tt large} means that the model takes more parameters. Due to potential privacy concerns, we rename several models, including hf\_public\_text\_generator1(hf\_ptg1), hf\_public\_text\_generator1\_large(hf\_ptg1\_large), public\_image\_generator1(pig1), and public\_image\_generator2(pig2).

\paragraph{Computer Vision} Computer vision is one of the most important domains that embrace deep learning. We further categorize models in different subareas. 
\begin{itemize}[leftmargin=*]
\item \textit{Image classification}, which categorizes and labels images. \tool{} includes 20 models in this domain, such as ResNet and its variant, MobileNet\_v2, and various models from the model collection timm~\cite{rw2019timm}. 

\item \textit{Object detection}, which detects all instances of predefined classes and provides axis-aligned boxes to locate the detected objects. \tool{} includes 12 models in this domain, including FasterRCNN and MaskRCNN atop Detectron2 and many other models. 

\item \textit{Image generation}, which takes texts or images as inputs and generates new images. \tool{} includes
pig2~\cite{dalle2}, a state-of-the-art diffusion model that can create realistic images and art from a description in natural language, and multiple GAN~\cite{gan} based models such as pig1~\cite{choi2018stargan} and CycleGAN~\cite{CycleGAN2017,isola2017image}. 

\sloppy
\item \textit{Image segmentation}, which partitions an image into multiple segments to locate objects and their boundaries. \tool{} includes YoLOv3~\cite{redmon2018yolov3} and a series of MaskRCNN models~\cite{he2017mask} implemented atop Detectron2. 

\item \textit{Pattern Recognition and Video Interpolation}. \tool{} includes background matting~\cite{sengupta2020background} in pattern recognition, which separates the foreground elements from the background of an image or video and composites it into a new background. \tool{} also includes Super SloMo~\cite{paliwal2020deep}, which reconstructs high-resolution slow-motion videos by alignment and appearance estimation.
\end{itemize}

\paragraph{Natural Language Process (NLP)} NLP is a series of algorithms and techniques enabling computers to understand human language. \tool{} includes NLP models for the task of language modeling and translation.
\begin{itemize}[leftmargin=*]
    \item \textit{Language Modeling}, which predicts the probability distribution of words in a language and models the relationships and dependencies among the words. \tool{} includes the open-sourced hf\_ptg1~\cite{gpt2} from Hugging Face platform and hf\_ptg1\_large with more parameters. Other models such as Bart~\cite{lewis2019bart}, Bert\cite{devlin2018bert}, T5~\cite{t5}, BigBird~\cite{zaheer2020big} are included to cover NLP domains such as natural language generation, translation, and comprehension.
    \sloppy
    \item \textit{Translation}, which converts text written in one language into text in another language while preserving its meaning as much as possible. 
    \tool{} includes the  transformer model attention\_is\_all\_you\_need\_pytorch, which is from a classic paper~\cite{vaswani2017attention}. 
\end{itemize}

\sloppy
\paragraph{Recommendation} Recommendation systems are used to give suggestions on personalized videos, advertisements, and text content.  \tool{} includes DLRM~\cite{DLRM19} open-sourced by Facebook Research which combines dense and sparse features to represent user-item interactions and generates recommendations based on these representations.
\tool{} also includes a PyTorch implementation of NVIDIA's DeepRecommender~\cite{deeprecommender}, which is based on a deep autoencoder with six layers and is trained end-to-end without any layer-wise pre-training.

\paragraph{Reinforcement Learning} Reinforcement learning focuses on the capability of making decisions from unstructured input data without manual engineering of the state space. \tool{} includes three representative models, drq~\cite{drq}, soft actor critic~\cite{haarnoja2018soft}, and learning to paint~\cite{huang2019learning}. 

\paragraph{Speech} The speech domain focuses on text and audio transformation and augmentation, such as 
speech recognition, synthesis, and audio source separation. \tool{} includes four models, speech\_transformer~\cite{li2019speechtransformer}, tactron2~\cite{shen2018natural} from NVIDIA,  tts\_angular~\cite{tacotron2} from Mozilla, and demucs\cite{demucs}.

\paragraph{Others}

\tool{} includes ten models from miscellaneous domains, such as pyhpc\_isoneutral\_mixing~\cite{pyhpc_isoneutral_mixing} from  high-performance computing and pytorch\_struct\cite{alex2020torchstruct} from core structured prediction algorithms for deep learning applications.

\input{code_computation.tex}

\subsection{Benchmark Adaptation and Configuration}

We adapt the off-the-shelf models to fulfill the requirement of benchmarks. We configure the computation of each model to exclude the data loading and postprocessing phases. We also select the optimal batch size and configure the precision to obtain a fair performance analysis.

\paragraph{Computation configuration} To focus on computation, TorchBench does not run the original end-to-end model code. Instead, it slices the original model code to run on a single GPU and only keeps the \textit{computation intensive} part, where most of the GPU computation happens. Listing~\ref{lst:e2e-train} shows the original model code of a ResNet50 model performing an image classification training task and highlights the computation-intensive part sliced by TorchBench. The rest parts, such as distributed setup, data preprocessing, model checkpointing, and data loading, are out of scope. 
Specifically, at the beginning of every model's benchmark test, we always assume that the input data has already been preprocessed and prefetched to the GPU device. To further simplify our analysis, we limit the TorchBench tests to run only 1 iteration repeatedly in the model code. We run each model ten times and report performance statistics of the run with the medium execution time.

\paragraph{Batch size configuration} Besides limiting the number of iterations, we also configure the model inputs by carefully specifying \texttt{batch\_size}, an important parameter in many deep learning tasks that defines the size of input per iteration. After data preprocessing, the input data is organized as a list of batches, where each batch corresponds to \texttt{batch\_size} number of the original data samples. The lower bound for \texttt{batch\_size} is 1, and the upper bound is limited by the capacity of GPU memory. For training, we use the default \texttt{batch\_size} value in the original model code, because \texttt{batch\_size} in training could affect model convergence. For inference, the original model code usually does not specify an optimal \texttt{batch\_size} of inputs. Thus, we run a set of tests with exhaustively enumerating \texttt{batch\_size} values (i.e., starting with one and doubling the size in each test) to determine the optimal value that yields the highest GPU utilization.
We believe our configuration enables optimal performance for each model in TorchBench.


\begin{figure*}[th]
    \centering
    \includegraphics[width=0.98\linewidth]{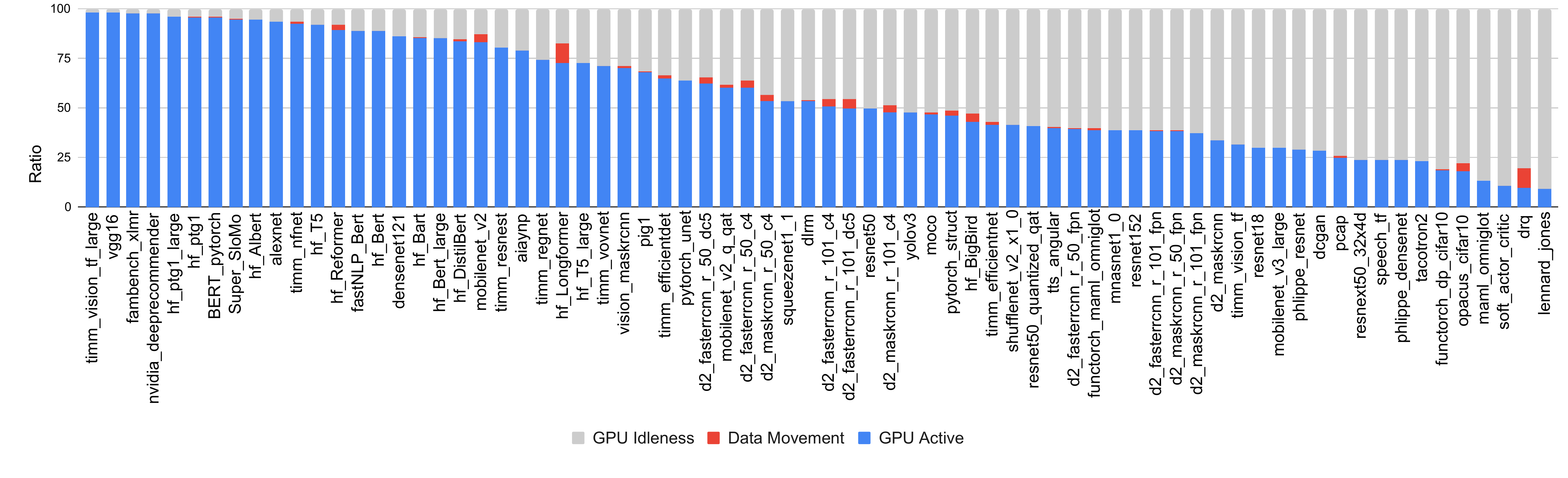}
    \caption{The execution time breakdown for training models in TorchBench.}
    \label{fig:missing_tflops_train}
\end{figure*}

\begin{figure*}[th]
    \centering
    \includegraphics[width=\linewidth]{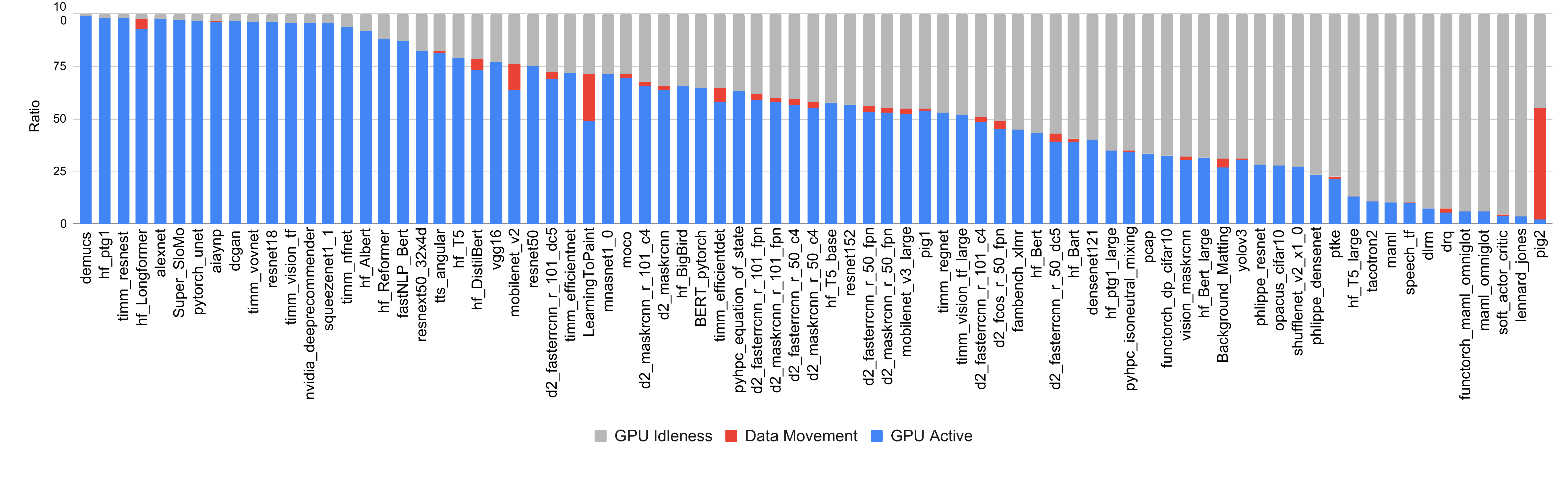}
    \caption{The execution time breakdown for inference models in TorchBench.} 
    \label{fig:missing_tflops_infer}
\end{figure*}

\paragraph{Precision configuration} We use the 32-bit floating point operations (FP32 or TF32) for all the benchmarks unless models have specific requirements. Although TorchBench supports other data precisions such as FP16~(half-precision), BF16~(Brain Floating Point Format), and AMP~(automatic mixed precision)~\cite{amp}, FP32 and TF32 are the most representative and recommended precisions. On the one hand, GPU vendors such as NVIDIA publish concrete roofline performance numbers for FP32 operations in TFLOPS~(Tera Floating Point Operations per Second) as part of the hardware specification, which facilitates our characterization efforts. For example, the theoretical peaks of an NVIDIA Tesla A100 GPU and an AMD Instinct MI210 GPU are 19.5 TFLOPS~\cite{a100} and 22.6 TFLOPS~\cite{mi210}, respectively. 
On the other hand, PyTorch uses TF32 for cuDNN by default, as TF32 is newly developed and typically yields better performance than FP32.

\subsection{\tool{} vs. MLPerf}

The goals of designing \tool{} and MLPerf are different. \tool{} aims to give a comprehensive and deep analysis of PyTorch software stack, while MLPerf aims to compare models running atop different frameworks. Thus,
\tool{} differs from MLPerf in three aspects.
\begin{itemize}[leftmargin=*]
    \item \tool{} benchmarks the computation phase of DL models, while MLPerf benchmarks the end-to-end execution of the models. 
    
    \item \tool{} benchmarks PyTorch only, while MLPerf benchmarks different deep learning frameworks.
    
    \item \tool{} includes 84 DL models in six domains, while MLPerf has only five models in five domains with PyTorch. \tool{} covers 2.3$\times$ more PyTorch APIs than MLPerf.
\end{itemize}
Recently, \tool{} has evolved to embrace end-to-end models and support beyond PyTorch (e.g., Jax). However, this evolution is in the preliminary stage and out of the scope of this paper.

\section{\tool{} Chracterization}
\label{characterization}

\tool{} enables comprehensive characterization of PyTorch. Given the page limit, we show the insights obtained from three characterization efforts: (1) characterizing PyTorch performance on NVIDIA GPUs (Section~\ref{section: characterize PyTorch Compute}), (2) characterizing PyTorch performance for different compiler backends (Section~\ref{section: pytorch compilers}), and (3) comparing PyTorch performance between NVIDIA and AMD GPUs (Section~\ref{section: amd vs nvidia}).

We benchmark \tool{} with PyTorch 2.0-20230102 nightly release linked with CUDA 11.7~\cite{pytorch}. 
Our experiments are done on one NVIDIA A100 GPU with 40 GB memory. Experiments in Section~\ref{section: amd vs nvidia} also include data obtained from an AMD MI210 GPU with 64 GB memory.

\input{table_category_active}

\subsection{Characterizing PyTorch Computation on GPU}\label{section: characterize PyTorch Compute}
We choose execution time as our main metric for GPU utilization because it is the most straightforward and common metric to measure the model performance. 

Figures~\ref{fig:missing_tflops_train} and~\ref{fig:missing_tflops_infer} show the characterization results for the training and inference tasks of TorchBench models, respectively. 
Each bar in the figures is composed of three segments: blue for the time that GPU is active for computation, red for the time used in data movement between CPU and GPU, and grey for the time that GPU is idle. We normalize them as portions of the total execution time of the models.
From the figures, we observe that PyTorch keeps GPU busy for only 56.8\% and 55.4\% of total execution time for training and inference, respectively. GPU idleness and CPU-GPU data movement account for a substantial time portion, preventing PyTorch from achieving full GPU usage. 
Table~\ref{table:category_active} further quantifies the time decomposition averaged across models in different domains. We obtain the following insights for further investigation.

\begin{figure*}[t]
    \centering
    \includegraphics[width=1\linewidth]{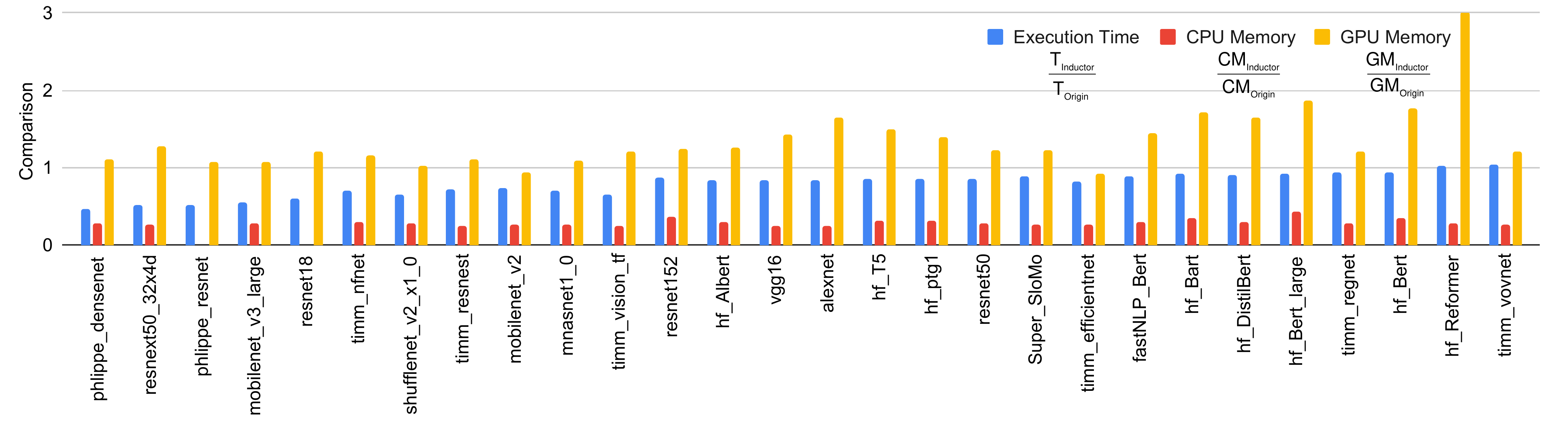}
    \caption{The comparisons of execution time (T), CPU memory usage (CM), and GPU memory usage (GM) for training between original PyTorch and PyTorch compiled by TorchInductor. $<1$ means TorchInductor performs better, while $>1$ means original PyTorch compiler performs better.}
    \label{fig:compiler_inductor1}
\end{figure*}
\begin{figure*}[t]
    \centering
    \includegraphics[width=1\linewidth]{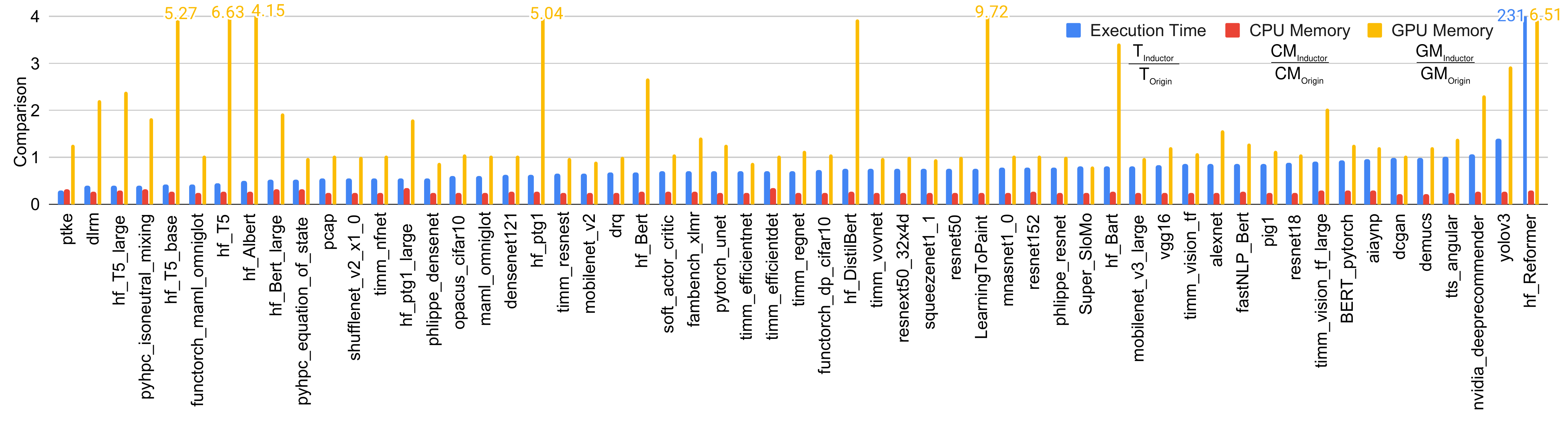}
    \caption{The comparisons of execution time (T), CPU memory usage (CM), and GPU memory usage (GM) for inference between original PyTorch and PyTorch compiled by TorchInductor. $<1$ means TorchInductor performs better, while $>1$ means original PyTorch compiler performs better.}
    \label{fig:compiler_inductor2}
\end{figure*}

\textit{Insights for execution time decomposition}
For both training and inference, models in computer vision, NLP, and recommendation yield over 50\% GPU active time, among which NLP models achieve $>$80\% in training. Models in these domains usually have large input sizes and intensive computation to be offloaded to GPU. 
In contrast, Reinforcement Learning (RL) models achieve the smallest GPU active time for both training and inference. RL models need to interact with {\tt environment}, a component not based on PyTorch, which limits the parallelism of RL models. Compared with NLP models, 
RL models have smaller inputs and less intensive computations in each batch, so they usually incur more GPU idleness.

\textit{Insights for performance difference between training and inference}
Some models perform better on training, while others perform better on inference. There are three reasons. First, training may use different input sizes from inference, so PyTorch invokes different GPU kernels. Second, some functions in training require higher precision than inference. Third, PyTorch may invoke different GPU kernels for training and inference even when they have the same input. For example, the GPU active time of fambench\_xlmr on training is 98.0\% but only 44.7\% for inference with the same input. With further investigation, we found that this model uses FP32 for training but FP16 for inference by default. For the same forward phase, FP16 GPU kernels run faster, so GPU finishes its computation earlier, resulting in a larger portion of idleness.

\textit{Discussion for individual models} pig2 is one of the outliers, which spends 52\% of execution time on data movement. With further investigation, we find that in order to save GPU memory usage, it always keeps one neural network structure on GPUs and offloads all other structures to CPUs. After the computations, it copies all structures back to GPUs. This kind of ping-pong data movement wastes a lot of time.

These insights motivate the necessity of optimizing PyTorch to remefy the performance losses caused by GPU idleness and data movement. We elaborate our optimization efforts with the help of \tool{} in Section~\ref{section: pytorch optimization}. It is worth noting that high GPU active time does not mean no room for performance improvement. For example, the GPU active ratio for vgg16 is 98.3\% while its achieved TFLOPS is about 10.07, which still has a gap from the peak performance. This is caused by GPU kernel inefficiencies, such as device memory access delays, instruction dependency delays, shared memory bank conflicts, and many pipeline stall reasons. Further characterization of TFLOPS is out of the scope due to the page limit.

\subsection{Characterizing PyTorch Compilers}\label{section: pytorch compilers}

Besides the default model interpreter in the eager mode,
PyTorch provides multiple model compilers (also known as model backends) in graph modes, such as TorchScript~\cite{torchscript}, Torch-TensorRT~\cite{torch-tensorrt}, TorchDynamo~\cite{torchdynamo}, and TorchInductor~\cite{torchinductor}. TorchScript is a classic Just-In-Time (JIT) model compiler that traces and optimizes model code. Torch-TensorRT is an Ahead-of-Time (AOT) compiler, which utilizes NVIDIA TensorRT~\cite{nvidia-tensorrt} deep learning optimizer and runtime to compile models before deployment. TorchDynamo compiles arbitrary Python code into graphs, which can be further compiled. TorchInductor compiles the graphs generated by TorchDynamo into optimized C++/Triton kernels. The combination of TorchDynamo and TorchInductor is the latest and recommended JIT compiler for PyTorch 2.0~\cite{pt2.0}. We use TorchInductor to denote this combination in this paper.

Figures~\ref{fig:compiler_inductor1} and~\ref{fig:compiler_inductor2} compare the performance between TorchInductor and the default PyTorch compiler. We measure three metrics: execution time, CPU memory consumption, and GPU memory consumption. It is worth noting that we do not show every benchmark in \tool{} because TorchInductor is still in its early stage and does not fully support PyTorch APIs. From the figures, we observe that TorchInductor typically improves the execution time over the default compiler, with 1.30$\times$ and 1.46$\times$ speedups for training and inference on average (geomean). Moreover, TorchInductor significantly reduces the CPU memory consumption by 71.2\% and 73.7\% for training and inference but generally increases the demand for GPU memory by 31.2\% and 51.1\% for training and inference. Specifically, many models suffer from such GPU memory bloat as high as $>5\times$ compared to the default PyTorch compiler. 

TorchInductor obtains speedups mainly from three techniques. First, it fuses GPU kernels and utilizes Triton to generate faster kernels. For example, fusing two subsequent functions can eliminate intermediate computations, memory load and store operations. 
Second, it reorders the resource-intensive nodes in the graph to make the tradeoff between performance and resource usage. 
Third, it determines which buffers can be reused and when the node should be executed according to the data dependencies. 
To apply these techniques, TorchInductor has to use its memory cache allocators and record graph details, which results in larger GPU memory footprints. Since many intermediate operations have been removed, the CPU memory usage is reduced significantly. We have reported the large GPU memory usage to TorchInductor developers, who confirm this perform issue and promise to fix it in the next release.


\paragraph{Outlier discussion} Inferencing {\tt yolov3} and {\tt hf\_Reformer} shows significant slowdown with TorchInductor. It is because of the high just-in-time (JIT) compilation overhead introduced by TorchInductor. For most models, TorchInductor only needs to JIT compile once and use the jitted model in the following iterations. However, models such as {\tt hf\_Reformer} incur many guard checks in TorchInductor, which guarantee the correct execution but cause a high overhead. For example, {\tt hf\_Reformer} incurs 2699 guard checks, and 30\% are heavy guard checks such as dictionary keys check. We confirm this performance issue with TorchInductor developers. 

\paragraph{Insights} TorchInductor typically accelerates both training and inference models with consuming more GPU memory compared to the default compiler. However, TorchInductor is not suitable for models running on GPUs with limited memory unless applying further configurations, such as changing batch sizes or using quantization.

\subsection{PyTorch on NVIDIA vs. AMD GPUs}\label{section: amd vs nvidia}

\input{table_gpu_specs}

\begin{figure*}[th]
    \centering
    \includegraphics[width=\linewidth]{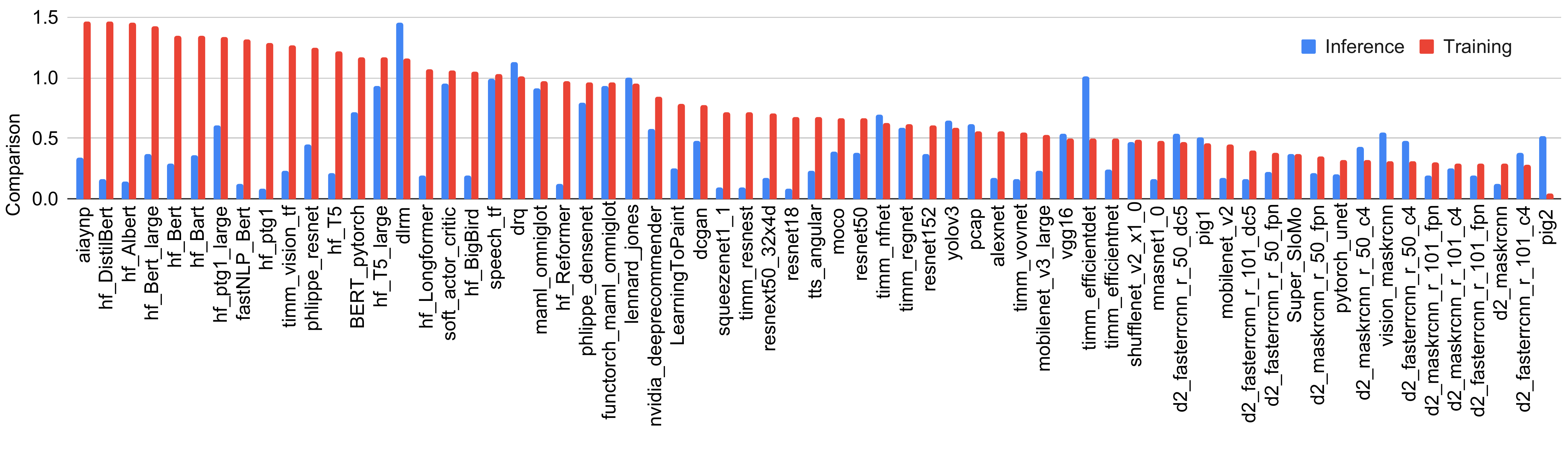}
    \caption{Comparing the execution time for training and inference obtained from NVIDIA A100 and AMD MI210 GPUs. Each bar represents the ratio as $\nicefrac{T_{NVIDIA}}{T_{AMD}}$. Note that $<$1 means A100 performs better, while $>$1 means MI210 performs better.}
    \label{fig:rocm_cuda}
\end{figure*}

PyTorch supports various types of GPUs. In this section, we compare the performance of NVIDIA A100 and AMD MI210, which are the competing products in the market. We compare their performance on \tool{} with their software stacks: ROCm 5.4.2~\cite{rocm} from AMD and CUDA 11.8~\cite{cuda} from NVIDIA. We run the default 32-bit configuration of \tool{} on both GPUs for a fair comparison. We test PyTorch stable 2.0.1 on both platforms.
Table~\ref{table:gpu spec} compares the peak theoretical TFLOPS for different floating point number formats on these two GPUs. Theoretically, MI210 has higher performance than A100 in FP32 and FP64 computation. However, both A100 and MI210 have unique features to give an uncertainty on \tool{} performance in comparison. For example, FP32-Matrix and FP64-Matrix are the unique optimized matrix operations for FP32 and FP64 on AMD GPUs, yielding high TFLOPS. TF32 is the unique 32-bit floating point format on A100, which yields high TFLOPS but with some accuracy losses; FP64-Tensor Core is the FP64 operations uniquely accelerated by NVIDIA Tensor Cores. 

Figure~\ref{fig:rocm_cuda} shows the comparison of execution time obtained from AMD MI210 ($T_{AMD}$) and NVIDIA A100 ($T_{NVIDIA}$). The ratio each bar represents in the figure is $\nicefrac{T_{NVIDIA}}{T_{AMD}}$; $<1$ means NVIDIA A100 performs better, and $>1$ means AMD MI210 performs better. Overall, we can find no GPU best for all \tool{} models. 
For model inference, AMD MI210 can achieve  1.46$\times$ performance than NVIDIA A100 on {\tt dlrm}. On the opposite, NVIDIA A100 can yield 12.57$\times$ speedup over AMD MI210 on {\tt hf\_ptg1}. A similar situation appears in the training phase as well. 

\paragraph{Insights}
Our further investigation shows that models typically benefit from A100 if most of their GPU kernels can use the TF32 format for computation because TF32 can achieve much higher TFLOPS than FP32 and FP32-Matrix according to Table~\ref{table:gpu spec}. However, not all models can use TF32, as TF32 incurs accuracy losses. For example, training most NLP models invokes \texttt{aten::matmul} operator, which requires the use of FP32 since PyTorch 1.12. Similar operators include \texttt{elementwise\_add}  and \texttt{elementwise\_div} with FP32 precision. In this case, AMD MI210 performs better because it has a higher TFLOPS than NVIDIA A100 on FP32.

\section{Applying \tool{} in Practice}\label{section: applying in practice}

We have applied \tool{} to guide performance optimization in the entire PyTorch software stack, including DL models, PyTorch framework, and GPU acceleration libraries. 
We describe two ways to use \tool{}. First, we analyze each model in \tool{} to understand and optimize the GPU idleness and data movement on the source code level. We have identified five performance issues and three optimization patches have been upstreamed to PyTorch or model repositories, and two optimization patches are confirmed and under discussion for upstreaming. Second, we have the continuous integration (CI) service of PyTorch repository to include \tool{}. We perform a daily sanity check on the performance regression for every nightly release. We have identified seven commits that incur unexpected slowdown to multiple \tool{} benchmarks. Among these problematic commits, five are reverted and two are merged with optimization. In the remaining section, we elaborate on the use cases of both usages. 

\subsection{PyTorch Optimization with \tool{}} \label{section: pytorch optimization}

We optimize the entire PyTorch software stack to improve GPU utilization, characterized in Section~\ref{section: characterize PyTorch Compute}. We use PyTorch Profiler~\cite{pytorch_profiler} to understand GPU idleness and data movement. 

\input{code_zerograd}

\subsubsection{Minimizing GPU Idleness}
A GPU is said to be idle when there is no work scheduled on it. This is the grossest of inefficiencies because it means the precious GPU computation resource is wasted. As shown in Figures~\ref{fig:missing_tflops_train} and~\ref{fig:missing_tflops_infer}, \tool{} exposes significant GPU idleness in a majority of models. With the help of PyTorch Profiler, we are able to pinpoint the GPU idleness in both model and framework layers of the PyTorch software stack.

Listing~\ref{lst:zero_grad} shows an example in {\tt zero\_grad} method in PyTorch optimizer, which sets all gradients to zeros in each training iteration. In this method, \texttt{p.grad.zero\_} is invoked in a loop nest to set zeros for each gradient, which incurs a series of tiny GPU kernels. A significant amount of GPU idleness occurs in between these kernels. This inefficiency has been confirmed by PyTorch developers. We propose a fix as shown in Listing~\ref{lst:zero_grad}. 
We create a temporary list to maintain the references to all the gradients and utilize \texttt{torch.\_foreach\_zero\_} to set them to zeros with one GPU kernel. This optimization avoids GPU idleness due to waiting for kernel launches. Since the optimization is for the PyTorch framework, multiple models benefit from this optimization.

Moreover, we investigate \tool{} models that incur significant GPU idleness. For example, hf\_BigBird~\cite{zaheer2020big}, a recently proposed transformer-based model, has more than 50\% GPU idleness. We find that substantial computation is done on CPU, rather than GPU, resulting in the high GPU idleness. Our optimization is either to reduce the CPU execution time to reduce the GPU waiting time or offload some CPU work to GPU to keep GPU busy. 

\input{code_hf_reformer}

\subsubsection{Reducing Data Movement}\label{subsection:data_movement}
Excessive data movement between CPUs and GPUs is a well-known bottleneck in GPU applications. Some \tool{} models show nontrivial data movement. For example, training hf\_reformer incurs 2.9\% of execution time for data movement. 
Listing~\ref{lst:hf-reformer} shows the problematic method, {\tt \_len\_and\_dim\_norm}. This method calls \texttt{torch.rsqrt()}, which takes a tensor as the input to compute the reciprocal of square root of scalar \texttt{self.attention\_head\_size}. As \texttt{torch.rsqrt()} launches a GPU kernel for the computation, it needs to copy this scalar from CPU to GPU, which incurs a large overhead surpassing the benefit from GPU acceleration.  

Thus, we use \texttt{numpy.sqrt()} instead to perform the computation on the CPU. PyTorch then generates one single kernel for a division between the tensor and the scalar.
This optimization yields a 27$\times$ speedup for function \texttt{\_len\_and\_dim\_norm}. This optimization patch has been upstreamed to the PyTorch Transformer repository~\cite{git_transformer}. 

Moreover, we found pig2 spends 52.7\% of execution time on CPU-GPU memory copies in inference. As described in Section~\ref{section: characterize PyTorch Compute}, to save GPU memory usage, it keeps neural network structure on GPUs onlyand offloads all other neural network structures to CPUs and copy them back to GPUs upon needs. 
However, for GPUs with large memory capacity like NVIDIA A100, the offloads waste a lot of time. After a discussion with the pig2 developer, they add an option to disable such memory offloading, which yields a 10.1$\times$ speedup. 

\begin{figure}[t]
    \centering
    \includegraphics[width=\linewidth]{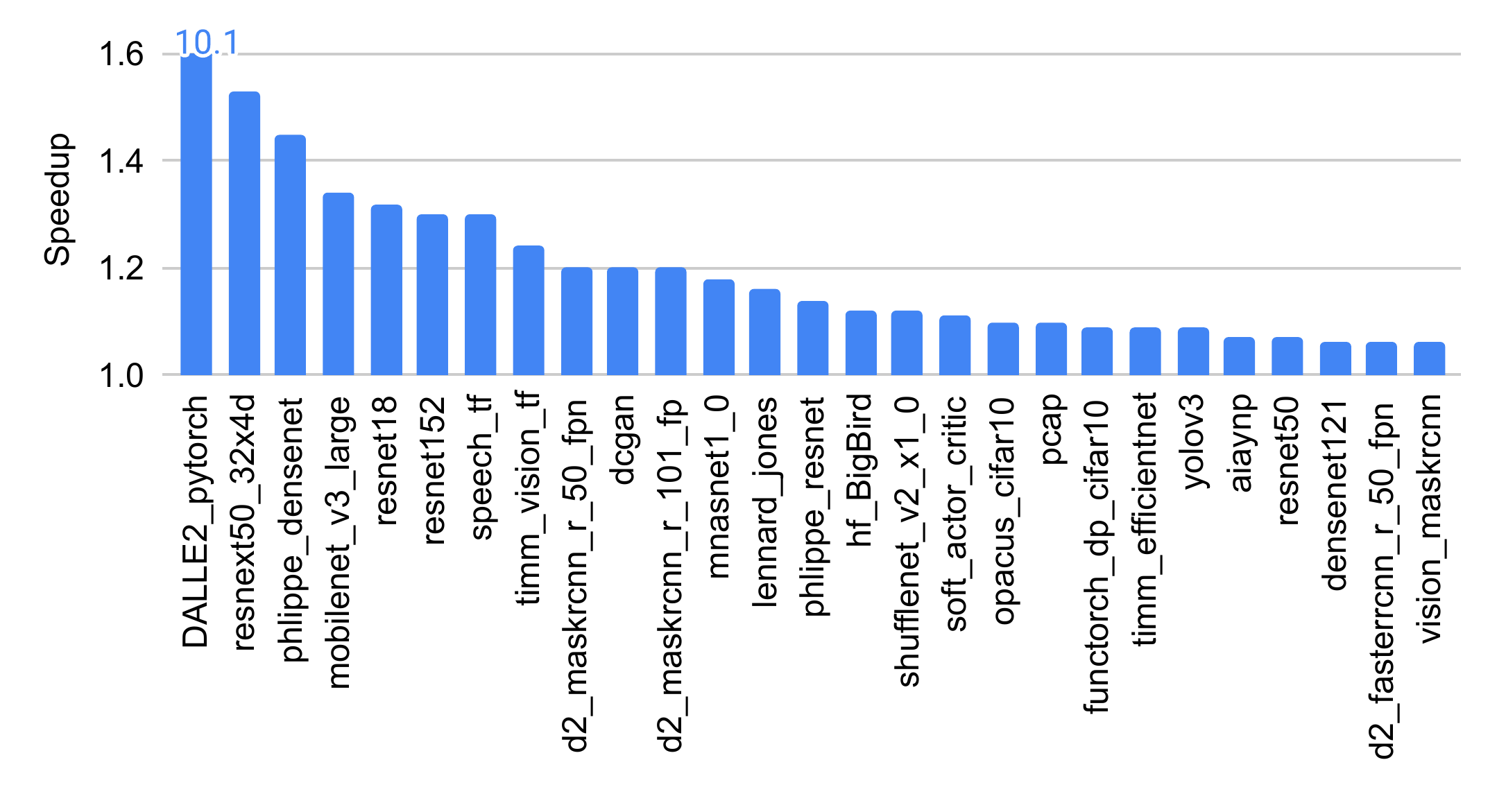}
    \caption{We only show models having speedups over 5\% obtained by applying different optimizations in the training phase. Other models do not show obvious performance changes due to our optimizations.}
    \label{fig:opt_speedup}
\end{figure}

\subsubsection{Optimization Speedups}
We run each \tool{} model 20 times and report the arithmetic mean of the obtained speedups with our optimization. 
In summary, 41 models out of 84 models yield nontrivial speedups in the training phase, which is 1.34$\times$ on average and up to 10.1$\times$. 
Our optimizations yield speedups over 1.03$\times$ for 15 out of 84 modes for inferences, with up to 10.3$\times$ speedup for pig2. The rest models do not show obvious improvement because our optimizations do not impact their execution paths.

It is worth noting that the PyTorch eco-system has already been thoroughly optimized for high performance by machine learning engineers and PyTorch system engineers. \tool{} can still help improve the performance significantly, which proves the value of \tool{}.
 

\subsection{PyTorch CI with \tool{}}

Continuous integration (CI) of a project repository enables automating the integration of code commits from multiple contributors. PyTorch leverages GitHub CI service
for pull request sanity checks, workflow checks, programmatic and stylistic error checks, OS and compiler checks, and many others. However, PyTorch lacks continuous performance checks.
We integrate \tool{} into the PyTorch CI service for performance regression testing. This is the first such effort for the PyTorch repository. In this section, we first describe how we set up PyTorch CI with \tool{} and then show the optimization for performance regression due to various reasons.

\subsubsection{PyTorch CI Setup with \tool{}}

Based on GitHub actions, we create a series of GitHub workflows to continuously test the performance regression with \tool{}. We measure the following two metrics.

\begin{itemize}[leftmargin=*]
    \item {\em Execution time}. \tool{} can be configured to run on CPU only or CPU+GPU for model training or inference. CI measures the execution time of each \tool{} benchmark in all four configurations: training on CPU, training on CPU+GPU, inference on CPU, and inference on CPU+GPU.
    
    \item {\em Memory usage}. For each \tool{} benchmark, CI measures the peak CPU and GPU memory consumption and checks the memory leaks. Similar to the time measurement, the memory measurement is also for all four configurations.
\end{itemize}

\input{table_all_issues.tex}

Naively, CI collects these metrics for every commit and checks whether a commit (also known as a pull request or PR) significantly bloat execution time or memory usage. From our experiences, we define the thresholds as a 7\% increment in execution time and memory usage. If at least one \tool{} benchmark exceeds the thresholds, PyTorch CI automatically submits a GitHub issue with the detailed performance report and the problematic commit for further investigation.

However, there is a large overhead for CI to perform these checks on each commit because more than 70 commits can be submitted every day. To reduce the overhead, PyTorch CI  performs the checks only on PyTorch's nightly version, which is automatically built from the latest commit at the end of the day. If any performance regressions are found, CI uses the binary search to check the commits submitted on the same day ordered by their submission timestamps. This optimization can significantly reduce the CI overhead.

\subsubsection{Performance Regression Case Studies}
With the help of \tool{}, PyTorch CI is able to identify many submitted commits with performance issues. 
Table~\ref{table:all_issues} overviews some problematic commits and their issues. All these issues are reported to the PyTorch team and fixed immediately with either a patch update or patch reversion. We elaborate on several typical case studies.

\paragraph{Runtime inflation}

\input{code_cpu_blas.tex}
\input{table_template_misoverload.tex}

Most problematic PRs fall into this category. Listing~\ref{lst:cpu_blas} shows PR \#65839, which results in 6.82$\times$ slowdown on average for training and 24.47$\times$ slowdown for inference. These slowdowns are observed across six models as shown in Table~\ref{table:template_misoverload}. This PR updates the C++ template match from the type {\tt scalar\_t} to {\tt opmath\_t} in the {\tt gemm} function. The \texttt{opmath\_t} is supposed to be compiled to faster oneDNN-accelerated functions. However, such template matching results in slower type conversions introduced by compilers. This performance issue has been confirmed by PyTorch developers, and the PR is reverted.

\input{code_distribution.tex}

Listing~\ref{lst:distribution} shows part of PR \#61056, which adds a validity check on {\tt value} in the \texttt{torch.distributions} API. This API generates parameterizable probability distributions. This PR incurs 11\% slowdown to model soft\_actor\_critic. With further investigation, we find that line~\ref{label:dist_line7} uses \texttt{valid.all} to check all values in the array, which incurs a high overhead. However, this validity check is unnecessary as line~\ref{label:dist_line8} does the same check in another way. We report our findings to PyTorch developers and this performance issue is fixed by removing the redundant validity checks.  

PR \#65594 introduces Conv-Bias-Relu fusion to fuse possible function combinations. However, it results in 21\% slowdown on average for nine models on NVIDIA M60 GPU with cuDNN 7.6.5. The fix is patched to bypass the optimizations on devices such as M60 GPU. PR \#72148 introduces optimized bias fusions but causes 7.8\% slowdown on model nvidia\_deeprecommender for some input sizes. The fix is to set the correct workspace size related to bias fusions. PR \#87855 has been introduced in Section~\ref{section:motivation}. PR \#71904 introduces extra bound checks resulting in a 14.9\% slowdown for model dlrm. The fix is to remove the bound checks.

\paragraph{Memory bloat}
PR \#85447 raises a red flag in the CI when measuring the memory consumption metric. It introduces a potential memory leak that part of the memory is not reclaimed after the computation. We investigate the PR and find that PyTorch uses a different memory management scheme to interact with cuBLAS. Instead of having cuBLAS manages its own memory, this PR allows PyTorch to preallocate memory for cuBLAS workspace automatically. However, this PR does not free the workspace memory. We report this finding to the PR authors, who add the memory reclamation \texttt{torch.\_C.\_cuda\_clearCublasWorkspaces()} to avoid the memory leak.

\section{Related Work}\label{section:related work}

There are many existing approaches~\cite{nsightcompute,nsightsystem,adhianto2010hpctoolkit,zhou2020gvprof,zhou2022valueexpert,amdprofiler,shende2006tau,scorep,hao2023drgpu} to characterizing, analyzing, and optimizing general GPU applications. 
In this section,
we review the most related solutions that analyze and optimize the performance (i.e., execution efficiency, not model accuracy/robustness) of deep learning frameworks and models. 

\subsection{Benchmarking Deep Learning Worloads}

MLPerf~\cite{mattson2020mlperf} is one of the most popular deep learning benchmark suites with contributors across major industrial stakeholders. MLPerf embraces eight models to benchmark end-to-end execution of training and inferences. MLPerf is used to evaluate the model performance with different deep learning frameworks running on different hardware. DAWNBench~\cite{coleman2017dawnbench} and AIBench~\cite{gao2019aibench} follow a similar design and goal with their end-to-end benchmarking.
Additionally,
microbenchmark suites such as Fathom~\cite{adolf2016fathom}, DeepBench~\cite{deepbench}, DNNMark~\cite{dong2017dnnmark}, and AIBench, include multiple computation kernels (aka operators) widely used in deep learning workloads. These microbenchmarks configure the operators with different inputs to understand how these operators behave on different hardware.

\tool{} has a completely different design goal from existing benchmark suites. \tool{} aims to expose the performance issues in PyTorch repository along with the project evolution. It obtains deep insights in PyTorch code bases, but not focuses on the characterization across different deep learning frameworks.

\subsection{Profiling Deep Learning Worloads}
There are many profilers~\cite{zhou2020gvprof,zhou2022valueexpert,adhianto2010hpctoolkit} that are able to understand performance inefficiencies in CPU-GPU applications, including deep learning frameworks. Since they are not specialized for deep learning applications, they require significant manual efforts and domain knowledge to understand inefficiencies and devise actionable optimization. 
Domain-specific profilers that target deep learning frameworks include PyTorch Profiler~\cite{pytorch_profiler}, Tensorflow Profiler~\cite{tensorflow_profiler}, DLPerf~\cite{dlperf}, MXNet Profiler~\cite{mxnet_profiler}, to name a few. These profilers pinpoint hotspots in both CPU and GPU computation kernels and associate them with deep learning operators. These profilers can be used together with \tool{} for better performance insights.

\subsection{Optimizing Deep Learning Workloads} 
To obtain bare-metal performance for deep learning models on GPUs, a number of optimization methods have been proposed. Those optimization methods focus on two directions. One is domain- or task-specific highly tuned implementations. For example, CUDA Deep Neural Network library (cuDNN)~\cite{chetlur2014cudnn} released by NVIDIA is a GPU-accelerated library of primitives for deep neural networks. It has been widely applied in different deep learning frameworks. Intel OneAPI~\cite{oneapi}, Google's XNNPACK~\cite{xnnpack}, and Android's NNAPI~\cite{nnapi} among others are of similar purposes. TorchScript~\cite{torchscript} and Grappler~\cite{Grappler} are two common Python frameworks to optimize deep learning graph passes by utilizing compiler optimization techniques such as constant folding. There are also some other optimization techniques such as pruning~\cite{pruning1,pruning2,pruning3}, quantization~\cite{quantization1, quantization2, quantization3} and operator fusion~\cite{fusion1,fusion2,fusion3,fusion4}.

The other direction is optimizing compilers. XLA (Accelerated Linear Algebra)~\cite{sabne2020xla} and TorchDynamo~\cite{torchdynamo} are two linear algebra code compilers for Tensorflow and PyTorch separately. They can accelerate deep learning models with potentially no source code changes.  JAX~\cite{jax2018github} is recently released by Google to provide composable transformations of deep learning models. 
TVM~\cite{chen2018tvm} is a deep learning compiler that can autotune the deep learning models for given hardware.
AITemplate~\cite{aitemplate} fuses different operations in various hierarchies to remove unnecessary intermediate computations and memory usage as much as possible. TASO~\cite{taso} reduces the strength of the computation by transforming it to an equivalence of higher efficiencies.

\tool{} complements these optimization techniques by
providing a platform to evaluate all these existing optimization techniques and identify new optimization opportunities. \tool{} provides unique insights to enable optimization for performant PyTorch code bases.


\section{Conclusions}\label{section:conclusion}

This paper describes \tool{}, which is the first comprehensive benchmark suite for PyTorch. We show the unique insights obtained from \tool{} benchmarking the entire PyTorch software stack running on mainstream NVIDIA and AMD GPUs. Moreover, we show the real use cases of applying \tool{} to guide code optimization and support regression testing. With the help of \tool{}, we are able to devise many optimization patches to PyTorch, and most of them are upstreamed to PyTorch official repository.

\bibliographystyle{ieeetr}

\bibliography{references}
\end{document}

%% file: table_all_models_4.tex
\begin{table*}[]
\caption{\tool{} consists of 84 models from various domains to benchmark PyTorch.}
\label{table:all_models_3}
\centering
\footnotesize

\begin{tabular}{lllll}
\hline
Domain                                                                      & Task                                                                            & \multicolumn{3}{l}{Model}                                                                                \\ \hline
\multirow{18}{*}{\begin{tabular}[c]{@{}l@{}}Computer\\ Vision\end{tabular}} & \multirow{7}{*}{\begin{tabular}[c]{@{}l@{}}Image\\ Classification\end{tabular}} & alexnet                                & phlippe\_resnet                    & squeezenet1\_1             \\ \cdashline{3-5} 
                                                                            &                                                                                 & densenet121                            & resnet152                          & timm\_efficientnet         \\ \cdashline{3-5} 
                                                                            &                                                                                 & mnasnet1\_0                            & resnet18                           & timm\_nfnet                \\ \cdashline{3-5} 
                                                                            &                                                                                 & mobilenet\_v2                          & resnet50                           & timm\_regnet               \\ \cdashline{3-5} 
                                                                            &                                                                                 & mobilenet\_v2\_quantized\_qat          & resnet50\_quantized\_qat           & timm\_resnest              \\ \cdashline{3-5} 
                                                                            &                                                                                 & mobilenet\_v3\_large                   & resnext50\_32x4d                   & vgg16                      \\ \cdashline{3-5} 
                                                                            &                                                                                 & phlippe\_densenet                      & shufflenet\_v2\_x1\_0              &                            \\ \cdashline{2-5} 
                                                                            & \multirow{4}{*}{\begin{tabular}[c]{@{}l@{}}Object\\ Detection\end{tabular}}     & detectron2\_fasterrcnn\_r\_101\_c4     & detectron2\_fasterrcnn\_r\_50\_dc5 & doctr\_reco\_predictor     \\ \cdashline{3-5} 
                                                                            &                                                                                 & detectron2\_fasterrcnn\_r\_101\_dc5    & detectron2\_fasterrcnn\_r\_50\_fpn & timm\_efficientdet         \\ \cdashline{3-5} 
                                                                            &                                                                                 & detectron2\_fasterrcnn\_r\_101\_fpn    & detectron2\_maskrcnn               & timm\_vovnet               \\ \cdashline{3-5} 
                                                                            &                                                                                 & detectron2\_fasterrcnn\_r\_50\_c4      & doctr\_det\_predictor              & vision\_maskrcnn           \\ \cdashline{2-5} 
                                                                            & \multirow{2}{*}{\begin{tabular}[c]{@{}l@{}}Image\\ Generation\end{tabular}}     & timm\_vision\_transformer\_large       & pytorch\_CycleGAN\_and\_pix2pix    & dcgan                      \\ \cdashline{3-5} 
                                                                            &                                                                                 & timm\_vision\_transformer              & public\_image\_generator1                   & public\_image\_generator2            \\ \cdashline{2-5} 
                                                                            & \multirow{3}{*}{\begin{tabular}[c]{@{}l@{}}Image\\ Segmentation\end{tabular}}   & detectron2\_maskrcnn\_r\_101\_c4       & detectron2\_maskrcnn\_r\_50\_fpn   & yolov3                     \\ \cdashline{3-5} 
                                                                            &                                                                                 & detectron2\_maskrcnn\_r\_101\_fpn      & detectron2\_fcos\_r\_50\_fpn       &                            \\ \cdashline{3-5} 
                                                                            &                                                                                 & detectron2\_maskrcnn\_r\_50\_c4        & pytorch\_unet                      &                            \\ \cdashline{2-5} 
                                                                            & \begin{tabular}[c]{@{}l@{}}Pattern\\ Recognition\end{tabular}                   & Background\_Matting                    &                                    &                            \\ \cdashline{2-5} 
                                                                            & \begin{tabular}[c]{@{}l@{}}Video\\ Interpolation\end{tabular}                   & Super\_SloMo                           &                                    &                            \\ \hline
\multirow{7}{*}{NLP}                                                        & \multirow{5}{*}{\begin{tabular}[c]{@{}l@{}}Language\\ Modeling\end{tabular}}    & BERT\_pytorch                          & hf\_Bert\_large                    & hf\_Longformer             \\ \cdashline{3-5} 
                                                                            &                                                                                 & fambench\_xlmr                         & hf\_BigBird                        & hf\_Reformer               \\ \cdashline{3-5} 
                                                                            &                                                                                 & hf\_Albert                             & hf\_DistilBert                     & hf\_T5                     \\ \cdashline{3-5} 
                                                                            &                                                                                 & hf\_Bart                               & hf\_public\_text\_generator1                          & hf\_T5\_base               \\ \cdashline{3-5} 
                                                                            &                                                                                 & hf\_Bert                               & hf\_public\_text\_generator1\_large                    & hf\_T5\_large              \\ \cdashline{2-5} 
                                                                            & Translation                                                                     & attention\_is\_all\_you\_need\_pytorch &                                    &                            \\ \cdashline{2-5} 
                                                                            & Other                                                                           & fastNLP\_Bert                          &                                    &                            \\ \hline
Recommendation                                                              & -                                                                               & nvidia\_deeprecommender                & dlrm                               &                            \\ \hline
\begin{tabular}[c]{@{}l@{}}Reinforcement \\ Learning\end{tabular}           & -                                                                               & drq                                    & soft\_actor\_critic                & LearningToPaint            \\ \hline
\multirow{3}{*}{Speech}                                                     & Recognition                                                                     & speech\_transformer                    &                                    &                            \\ \cdashline{2-5} 
                                                                            & \begin{tabular}[c]{@{}l@{}}Audio Source\\ Separation\end{tabular}               & demucs                                 &                                    &                            \\ \cdashline{2-5} 
                                                                            & Synthesis                                                                       & tacotron2                              & tts\_angular                       &                            \\ \hline
\multirow{4}{*}{Other}                                                      & \multirow{4}{*}{-}                                                              & pytorch\_struct                        & pyhpc\_turbulent\_kinetic\_energy  & pyhpc\_equation\_of\_state \\ \cdashline{3-5} 
                                                                            &                                                                                 & functorch\_dp\_cifar10                 & opacus\_cifar10                    & pyhpc\_isoneutral\_mixing  \\ \cdashline{3-5} 
                                                                            &                                                                                 & functorch\_maml\_omniglot              & moco                               & maml                       \\ \cdashline{3-5} 
                                                                            &                                                                                 & lennard\_jones                         & maml\_omniglot                     &                            \\ \hline
\end{tabular}
\end{table*}

%% file: code_computation.tex
\begin{figure}[tp]
\begin{lstlisting}[label=lst:e2e-train, escapechar=|, caption={An example of ResNet50 deep learning model training task. In our benchmark, we only measure the TFLOPS of the highlighted program segment and set both \texttt{num\_epochs} and \texttt{len(train\_dataloader)} to 1. },linebackgroundcolor={
    \ifnum\value{lstnumber}=10
            \color{blue!10}
    \fi
    \ifnum\value{lstnumber}=11
            \color{blue!10}
    \fi
    \ifnum\value{lstnumber}<19
    \ifnum\value{lstnumber}>13
            \color{blue!10}
    \fi
    \fi
}
]
def main():
    init_distributed()
    model = create_model("resnet50")
    train_dataloader = create_dataloader(TRAIN_DATASET_DIR)
    val_dataloader = create_dataloader(VAL_DATASET_DIR)
    optimizer = create_optimizer(model)
    lr_scheduler = create_scheduler(optimizer)
    loss_fn = torch.nn.Loss()
    # start training loop
    for epoch in range(start_epoch, num_epochs):
        for batch in train_dataloader:
            inputs, targets = batch
            inputs, targets = inputs.cuda(), targets.cuda()
            outputs = model(inputs)}
            loss = loss_fn(outputs, targets)
            optimizer.zero_grad()
            loss.backward()
            optimizer.step()
        validate(model, val_dataloader)
        lr_scheduler.step()
        save_checkpoint() # save model checkpoint after each epoch
\end{lstlisting}
\end{figure}

%% file: table_category_active.tex
\begin{table*}[]
\footnotesize
\caption{The breakdown ratios of model execution time for different deep learning tasks.}
\label{table:category_active}
\centering

\begin{tabular}{|c|ccc|ccc|}
\hline
\multirow{2}{*}{Task}  & \multicolumn{3}{c|}{Train}                                                      & \multicolumn{3}{c|}{Inference}                                                  \\ \cline{2-7} 
                       & \multicolumn{1}{c|}{GPU activeness} & \multicolumn{1}{c|}{Data movement} & \multicolumn{1}{c|}{GPU idleness} & \multicolumn{1}{c|}{GPU activeness} & \multicolumn{1}{c|}{Data movement} & \multicolumn{1}{c|}{GPU idleness} \\ \hline
Computer Vision        & \multicolumn{1}{c|}{53.1}       & \multicolumn{1}{c|}{2.1}           & 44.8     & \multicolumn{1}{c|}{62.8}       & \multicolumn{1}{c|}{1.4}           & 35.7     \\ \hline
NLP                    & \multicolumn{1}{c|}{84.9}       & \multicolumn{1}{c|}{1.3}           & 13.8     & \multicolumn{1}{c|}{64.7}       & \multicolumn{1}{c|}{0.8}           & 34.5     \\ \hline
Recommendation         & \multicolumn{1}{c|}{75.4}       & \multicolumn{1}{c|}{0.4}           & 24.2     & \multicolumn{1}{c|}{51.4}       & \multicolumn{1}{c|}{0.1}           & 48.5     \\ \hline
Reinforcement Learning & \multicolumn{1}{c|}{10.2}       & \multicolumn{1}{c|}{5.0}           & 84.8     & \multicolumn{1}{c|}{19.3}       & \multicolumn{1}{c|}{8.4}           & 72.3     \\ \hline
Speech                 & \multicolumn{1}{c|}{28.8}       & \multicolumn{1}{c|}{0.3}           & 70.9     & \multicolumn{1}{c|}{50.3}       & \multicolumn{1}{c|}{0.3}           & 49.4     \\ \hline
\end{tabular}
\end{table*}

%% file: table_gpu_specs.tex
\begin{table}[]
\scriptsize
\caption{The peak theoretical TFLOPS for various floating point number formats on NVIDIA A100 and AMD MI210 GPUs.}
\label{table:gpu spec}
\centering

\begin{tabular}{|c|cccccc|}
\hline
\multirow{2}{*}{GPU} & \multicolumn{6}{c|}{TFLOPS of Floating Point Number Formats}                                                                                                                                                                                                                                                            \\ \cline{2-7} 
                     & \multicolumn{1}{c|}{FP32} & \multicolumn{1}{c|}{TF32} & \multicolumn{1}{c|}{\begin{tabular}[c]{@{}c@{}}FP32-\\ Matrix\end{tabular}} & \multicolumn{1}{c|}{FP64} & \multicolumn{1}{c|}{\begin{tabular}[c]{@{}c@{}}FP64-\\ Matrix\end{tabular}} & \begin{tabular}[c]{@{}c@{}}FP64-\\ Tensor Core\end{tabular} \\ \hline
NVIDIA A100          & \multicolumn{1}{c|}{19.5} & \multicolumn{1}{c|}{156}  & \multicolumn{1}{c|}{-}                                                      & \multicolumn{1}{c|}{9.7}  & \multicolumn{1}{c|}{-}                                                      & 19.5                                                        \\ \hline
AMD MI210            & \multicolumn{1}{c|}{22.6} & \multicolumn{1}{c|}{-}    & \multicolumn{1}{c|}{45.3}                                                   & \multicolumn{1}{c|}{22.6} & \multicolumn{1}{c|}{45.3}                                                   & -                                                           \\ \hline
\end{tabular}
\end{table}

%% file: code_zerograd.tex
\begin{figure}[tp]
\begin{lstlisting}[label=lst:zero_grad, escapechar=|, caption={\texttt{zero\_grad} sets all gradients to zeros serially.},linebackgroundcolor={
    \ifnum\value{lstnumber}=4
            \color{red!10}
    \fi
    \ifnum\value{lstnumber}=16
            \color{blue!10}
    \fi
    \ifnum\value{lstnumber}=17
            \color{red!10}
    \fi
    \ifnum\value{lstnumber}<12
    \ifnum\value{lstnumber}>7
            \color{blue!10}
    \fi
    \fi
    \ifnum\value{lstnumber}<16
    \ifnum\value{lstnumber}>11
            \color{red!10}
    \fi
    \fi
}]   
def zero_grad():
  ...
  //pddg: per_device_and_dtype_grads
 +pddg = defaultdict(lambda: defaultdict(list))
  for group in self.param_groups:
    for p in group['params']:
      ...
 -    if (not foreach or p.grad.is_sparse):
 -      p.grad.zero_()
 -    else:
 -      pddg[p.grad.device][p.grad.dtype].append(p.grad)
 +    if not p.grad.is_sparse and p.grad.is_cuda:
 +      pddg[p.grad.device][p.grad.dtype].append(p)
 +    else:
 +      p.grad.zero_()
 -if foreach:
 +if foreach or pddg:
    for _, per_dtype_grads in pddg.items():
      for grads in per_dtype_grads.values():
        torch._foreach_zero_(grads)
\end{lstlisting}
\end{figure}

%% file: code_hf_reformer.tex
\begin{figure}[tp]
\begin{lstlisting}[label=lst:hf-reformer, escapechar=|, caption={Model hf\_reformer calls the function \texttt{torch.rsqrt()} to calculate the reciprocal of the square-root of the variable \texttt{self.attention\_head\_size}.},linebackgroundcolor={
    \ifnum\value{lstnumber}<7
    \ifnum\value{lstnumber}>2
            \color{blue!10}
    \fi
    \fi
}
]
def _len_and_dim_norm(self, vectors):
    vectors = self._len_norm(vectors)
    vectors = vectors * torch.rsqrt(|\label{code:rsqrt}|
        torch.tensor(self.attention_head_size, 
                device=vectors.device, dtype=vectors.dtype)
    )
    return vectors
\end{lstlisting}
\end{figure}

%% file: table_all_issues.tex
\begin{table}[t]
\footnotesize
\caption{Seven issues found in PyTorch development by \tool{}.}\label{table:all_issues}

\centering

\begin{tabular}{|c|c|c|c|}
\hline
PR\#   & Issues                                                                         & Performance Issues & Fixed          \\ \hline
85447 & Break-chain API change                                                        & Memory bloat      & $\checkmark$ \\ \hline
61056 & Duplicate error check                                                         & Runtime inflation & $\checkmark$ \\ \hline
65594 & \begin{tabular}[c]{@{}c@{}}Optimization's device\\ compatibility\end{tabular} & Runtime inflation & $\checkmark$ \\ \hline
72148 & \begin{tabular}[c]{@{}c@{}}Suboptimal library\\ configuration\end{tabular}                                                  & Runtime inflation & $\checkmark$ \\ \hline
71904 & Redundant bound checks                                                        & Runtime inflation & $\checkmark$ \\ \hline
65839 &Template Mismatch                                                     & Runtime inflation & Reverted     \\ \hline
87855 & Misused error handling                                                        & Runtime inflation & Reverted     \\ \hline
\end{tabular}
\end{table}

%% file: code_cpu_blas.tex
\begin{figure}[tp]
\begin{lstlisting}[label=lst:cpu_blas, escapechar=|, caption={PR \#65839 causes inefficient template matching. This commit replaces the original \texttt{scalar\_t} with \texttt{opmath\_t} but causes inefficient template match in argument deduction.},linebackgroundcolor={
    \ifnum\value{lstnumber}=2
            \color{blue!10}
    \fi
    \ifnum\value{lstnumber}=7
            \color{blue!10}
    \fi
    \ifnum\value{lstnumber}=11
            \color{blue!10}
    \fi
    \ifnum\value{lstnumber}=1
            \color{red!10}
    \fi
    \ifnum\value{lstnumber}=6
            \color{red!10}
    \fi
    \ifnum\value{lstnumber}=10
            \color{red!10}
    \fi
}
]
-template <typename scalar_t>
+template <typename scalar_t, typename opmath_t=at::opmath_type<scalar_t>>
void gemm(
  TransposeType transa, TransposeType transb,
  int64_t m, int64_t n, int64_t k,
- scalar_t alpha,
+ opmath_t alpha,
  const scalar_t *a, int64_t lda,
  const scalar_t *b, int64_t ldb,
- scalar_t beta,
+ opmath_t beta,
  scalar_t *c, int64_t ldc) 
\end{lstlisting}
\end{figure}

%% file: table_template_misoverload.tex
\begin{table}[]
\centering
\caption{Six models has 15.64$\times$ slowdown on average for CPU testing. The slowdown is up to 51.37$\times$. }
\label{table:template_misoverload}
\footnotesize

\begin{tabular}{|c|c|c|}
\hline
Mode                       & Model            & Slowdown      \\ \hline
\multirow{4}{*}{Train}     & pytorch\_stargan & 20.45$\times$ \\ \cline{2-3} 
                           & vision\_maskrcnn & 3.68$\times$  \\ \cline{2-3} 
                           & maml\_omniglot   & 1.96$\times$  \\ \cline{2-3} 
                           & timm\_regnet     & 1.19$\times$  \\ \hline
\multirow{4}{*}{Inference} & pytorch\_stargan & 51.37$\times$ \\ \cline{2-3} 
                           & demucs           & 43.22$\times$ \\ \cline{2-3} 
                           & vision\_maskrcnn & 2.11$\times$  \\ \cline{2-3} 
                           & mnasnet1\_0      & 1.16$\times$  \\ \hline
\end{tabular}
\end{table}

%% file: code_distribution.tex
\begin{figure}[tp]
\begin{lstlisting}[label=lst:distribution, escapechar=|, caption={PR \#61056: Lines~\ref{label:dist_line4}-\ref{label:dist_line7} are added to check the values' validities, which are redundant.},linebackgroundcolor={
    \ifnum\value{lstnumber}<8
    \ifnum\value{lstnumber}>3
            \color{blue!10}
    \fi
    \fi
}
]
class Distribution(object):
  def __init__(...):
    ...
+   value = getattr(self, param)|\label{label:dist_line4}|
+   valid = constraint.check(value)
+   if not valid.all():
+     raise ValueError(...)|\label{label:dist_line7}|
    if not constraint.check(getattr(self, param)).all():|\label{label:dist_line8}|
      raise ValueError("The parameter {} has invalid  values".format(param))
\end{lstlisting}
\end{figure}